\documentclass{article}
\usepackage{spconf,amsmath,graphicx}
\usepackage{graphicx}
\usepackage{amsmath}
\usepackage{amssymb}
\usepackage{booktabs}
\usepackage{enumitem}
\usepackage{bm}

\usepackage[table]{xcolor}
\definecolor{lightblue}{rgb}{0.4, 0.7, 0.9}
\usepackage{colortbl}  
\usepackage{arydshln}

\usepackage{multirow}
\usepackage{multicol}
\usepackage{tabularx}
\usepackage{siunitx}
\usepackage{algorithm}
\usepackage{algorithmic}
\usepackage{paralist}

\usepackage{colortbl}
\usepackage{multirow}
\usepackage{booktabs}

\usepackage[pagebackref,breaklinks,colorlinks,bookmarks=false]{hyperref}

\usepackage[capitalize]{cleveref}
\crefname{section}{Sec.}{Secs.}
\Crefname{section}{Section}{Sections}
\Crefname{table}{Table}{Tables}
\crefname{table}{Tab.}{Tabs.}

\title{Certainty and Uncertainty Guided Active Domain Adaptation}
\name{ }
\address{ }
\name{Bardia Safaei, Vibashan VS, and Vishal M. Patel}
\address{Dept. of Electrical and Computer Engineering, Johns Hopkins University, MD, USA\\
\texttt{\{bsafaei1, vvishnu2, vpatel36\}@jhu.edu}}
\begin{document}
%
\maketitle
%
\begin{abstract}
   Active Domain Adaptation (ADA) adapts models to target domains by selectively labeling a few target samples. Existing ADA methods prioritize uncertain samples but overlook confident ones, which often match ground-truth. We find that incorporating confident predictions into the labeled set before active sampling reduces the search space and improves adaptation. To address this, we propose a collaborative framework that labels uncertain samples while treating highly confident predictions as ground truth. Our method combines Gaussian Process-based Active Sampling (GPAS) for identifying uncertain samples and Pseudo-Label-based Certain Sampling (PLCS) for confident ones, progressively enhancing adaptation. PLCS refines the search space, and GPAS reduces the domain gap, boosting the proportion of confident samples. Extensive experiments on Office-Home and DomainNet show that our approach outperforms state-of-the-art ADA methods.
   \noindent
\end{abstract}

\section{Introduction}
\label{sec:intro}
Deep Neural Networks (DNNs) have significantly advanced computer vision tasks. However, their performance degrades on data different from the training set \cite{ben2010theory}. To address this limitation, Unsupervised Domain Adaptation (UDA) methods \cite{tzeng2017adversarial,hoffman2018cycada,bousmalis2017unsupervised} improve generalization from a labeled source to an unlabeled target domain but still lag behind fully-supervised models trained on target data. While full annotation of target data is impractical, selectively labeling a small subset with human expertise is feasible.

\begin{figure}[t!]
    \begin{center}
        \includegraphics[width=0.92\linewidth]{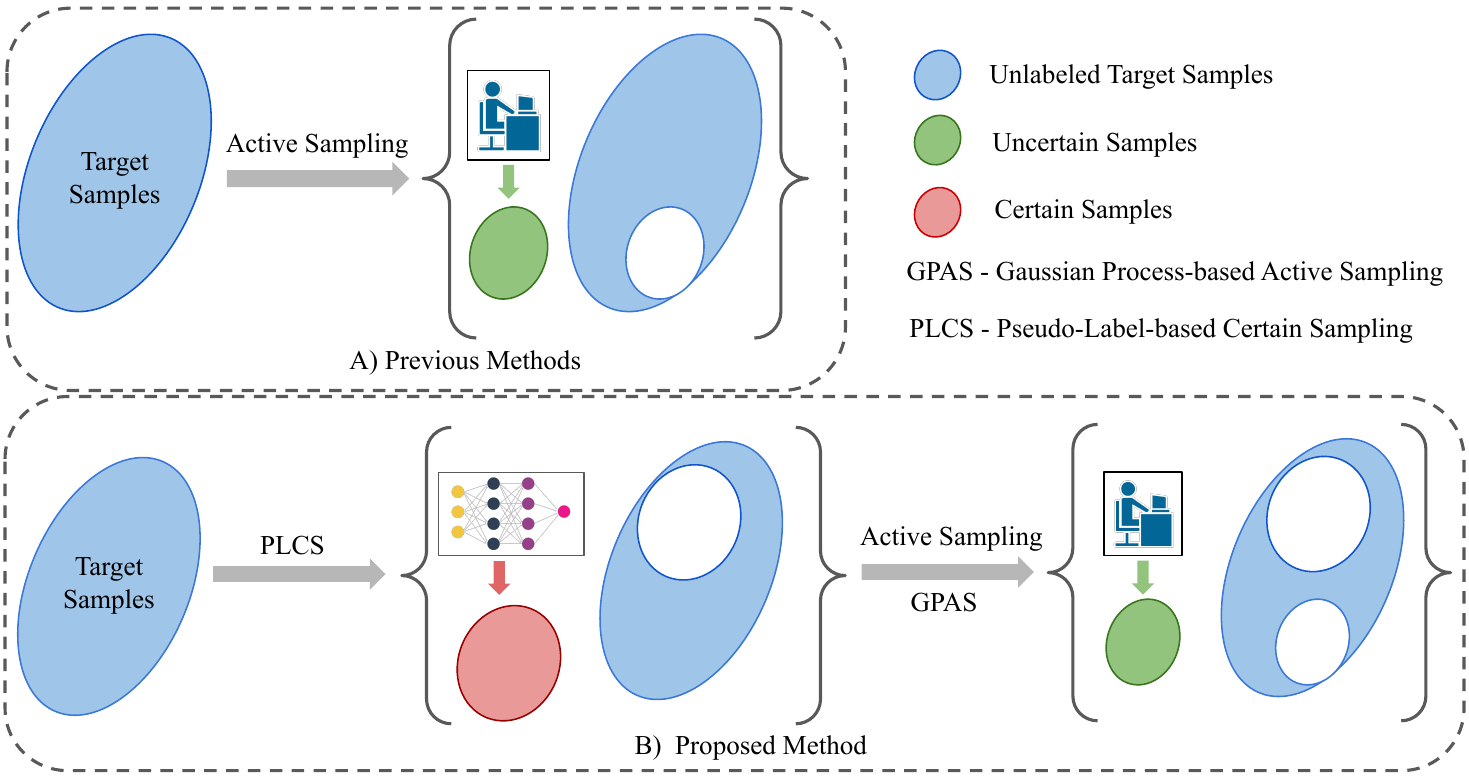}
    \end{center}
    \vskip -6.0mm
    \caption{\textbf{A) Previous Methods:} These methods focus on selecting uncertain/informative samples for labeling but overlook confident samples, missing valuable information. \textbf{B) Proposed Method:} Our work aims to address this limitation by integrating confident pseudo-labels alongside uncertain samples, enhancing active domain adaptation by leveraging both to improve model accuracy.}
    
    \label{fig:motivation} 
    \vskip -4.6mm
\end{figure}

Active Domain Adaptation (ADA) methods \cite{aada,sdm,tqs,xie2022active,prabhu2021active} improve model generalization to unlabeled target domains by annotating a few target samples during training. These annotations, combined with source data, enable more effective adaptation. These methods focus on selecting the most informative samples for annotation. For example, existing methods like CLUE \cite{prabhu2021active} and TQS \cite{tqs} select uncertain samples based on entropy and classifier disagreement, respectively. While effective, they rely solely on uncertainty and ignore confident samples (Fig. \ref{fig:motivation}). We argue that including confident samples into active sampling can lead to better and faster adaptation.

For domain adaptation, the source-only model’s most confident target samples often align with the ground truth. Incorporating these into the target labeled set before active sampling reduces the search space per round. In our experiments on DomainNet (sketch), progressively adding the top 30\% of confident samples during entropy-based active sampling reduced selection time from 601s to 540s (10\% faster). This efficiency is crucial for deploying ADA on large-scale datasets.

To this end, we propose a Gaussian Process-based Active Sampling (GPAS) strategy to identify uncertain/informative samples and a Pseudo-Label-based Certain Sampling (PLCS) strategy to identify highly confident samples, incorporating both into the labeled data. PLCS identifies highly confident target samples and integrates their label information into the target labeled set. To achieve this, we rank target samples by predicted class and select the top $\kappa\%$ confident samples from each class based on their pseudo-labels and confidence scores. These confident samples, which often match the ground truth, are directly added to the labeled data, reducing the active sampling search space before the active sampling process begins. GPAS aims to identify the most informative target samples using a probabilistic model called Gaussian Process (GP) \cite{rasmussen2006gaussian}. It constructs a Gaussian distribution for encoded features from both the source and target domains, creating a separate GP for each class to model their joint distribution. By combining all class posterior covariance matrices, GPAS estimates prediction uncertainty and selects the most uncertain samples with the highest variance for annotation, improving the model’s performance on the target domain. These two strategies work together in each sampling round, where PLCS reduces the active sampling search space and GPAS reduces the domain gap, boosting the rate of certain samples. Our paper makes the following contributions: i) We introduce a novel approach that improves ADA by integrating confident pseudo-labels into the sampling process without additional queries. ii) We propose two key strategies: GPAS for identifying uncertain samples and PLCS for selecting highly confident ones, enabling a more effective and efficient sampling process. iii) Extensive experiments on cross-domain benchmarks demonstrate that our method outperforms state-of-the-art ADA approaches.

\begin{figure*}[t!]
    \begin{center}
        \includegraphics[width=.88\linewidth]{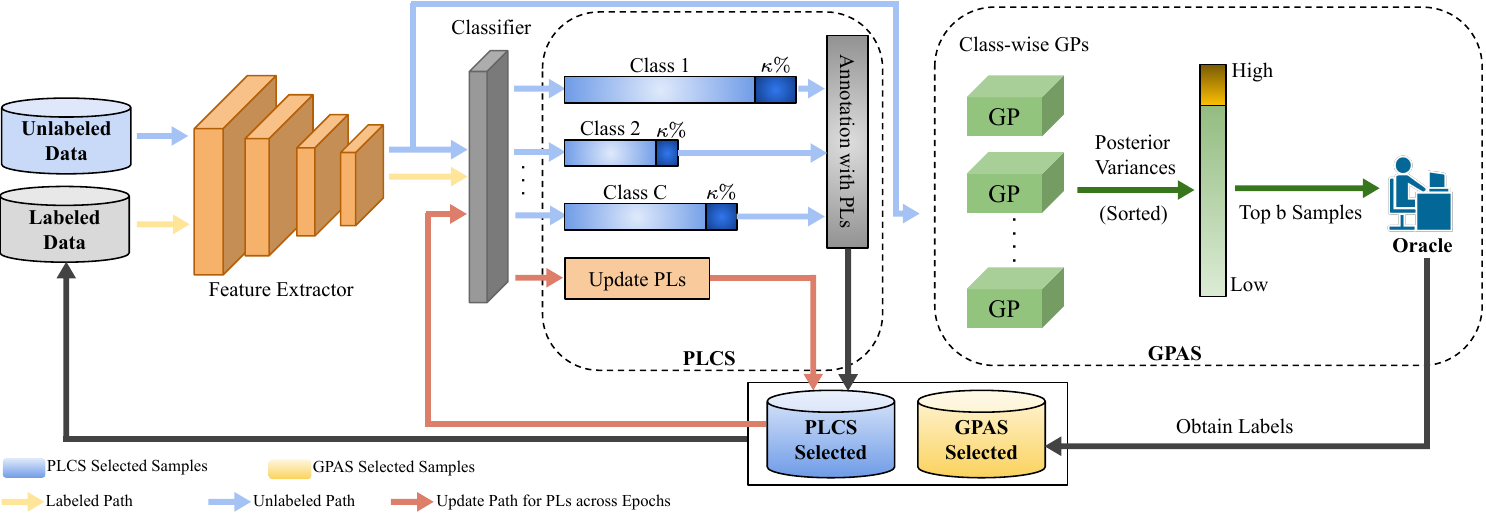}
    \end{center}
    \vskip -20.0pt
    \caption{Our approach consists of two complementary phases per sampling round. GPAS ranks unlabeled target samples by posterior variance using class-wise GPs and queries labels for the top $b$ samples, reducing domain shift and increasing confident PL selection. PLCS selects the top $\kappa\%$ most confident target samples per class, adding them to the labeled set with their pseudo-labels. This, in turn, helps GPAS by shrinking the query search space.}
    \label{fig:framework} 
    \vskip -5.0pt
\end{figure*}
\section{Background}
\textbf{Gaussian processes.} Gaussian Processes (GP) are non-parametric probabilistic models that generate uncertainty-aware predictions, making them suitable for semi-supervised and active learning \cite{yasarla2020syn2real,kapoor2007active}. A GP $f(x)$ is an infinite set of random variables where any finite subset follows a joint Gaussian distribution \cite{rasmussen2004gaussian}. It is defined as:  
\begin{equation}
 	f(x) \sim \mathcal{GP}(\mu(x), K(x, x')),
    \label{gp_prior}
\end{equation}
where $\mu(x)$ is the mean function and $K(x, x')$ is the kernel function. For simplicity, we assume $\mu(x) = 0$, though this is not required. The function values over a set of inputs follow a joint Gaussian distribution: 
\begin{equation}
\textbf{f} \sim \mathcal{N}(\bm{\mu},K(X,X')).
\vspace{-1.5mm}
\end{equation}
Using the GP prior in Eq. \ref{gp_prior}, we can compute the posterior GP to model the joint predictive distribution of labeled and unlabeled data, enabling more effective sample selection.

\medskip\noindent
\textbf{Active domain adaptation.} In ADA, we have a labeled source dataset $\mathcal{D}_{s}=\left\{ \left( x_{i}^{s},y_{i}^{s} \right) \right\}_{i=1}^{N_{s}}$ with $y_{i}^{s} \in \left\{ j\right\}_{j=1}^C$, and an unlabeled target dataset $\mathcal{D}_{t}=\left\{ \left( x_{i}^{t}\right) \right\}_{i=1}^{N_{t}}$, where both share the same label space but have different distributions. The target set is split into unlabeled data $\mathcal{D}_{ut}$ and labeled data $\mathcal{D}_{lt}$, which starts empty. Training begins with $\mathcal{D}_{s}$, and in each query round, a subset of $\mathcal{D}_{ut}$ is selected, labeled, and moved to $\mathcal{D}_{lt}$. Training continues on $\mathcal{D}_{l} = \mathcal{D}_{s} \cup \mathcal{D}_{lt}$ until the query budget $\mathcal{B}$ is exhausted. The goal is to maximize performance on target data by selecting the most informative samples.

\section{Approach}
\subsection{Pseudo-Label based Certain Sampling}
Existing Active Learning (AL) methods focus on selecting uncertain samples while ignoring confident ones. We improve this by leveraging pseudo-labels for confident target samples, reducing the need for active labeling and better approximating the target distribution. This makes AL more efficient by shrinking the search space and improving model accuracy. Formally, we assign each target sample a confidence score based on its highest predicted probability. In each round, we select the top $\kappa\%$ confident samples per class for supervised training, ensuring class diversity. As training progresses, $\kappa$ increases, and pseudo-labels are continuously updated based on the model’s latest predictions.
\subsection{Gaussian Process-based Active Sampling}
 To select the most informative unlabeled target samples in each active sampling round, we use GPs to estimate the posterior variance of each unlabeled sample in a class-wise manner. Specifically, GP serves as an indicator to quantify the uncertainty of target samples. We opt for class-wise GPs as modeling an unlabeled sample is computationally expensive, and inter-class uncertainty is difficult to model compared to intra-class uncertainty, as shown in Fig.~\ref{fig:framework}. For a given category $c$, let 
\begin{equation}
\begin{aligned}  
    X^u_{[c]} =\left\{ x_j^u \in \mathcal{D}_{ut}\,|\,\hat{y}_j=c \right\}_{j=1:N_u},\\
    X^l_{[c]} =\left\{ x_j^l\in \mathcal{D}_{l}\,|\,y_j=c \right\}_{j=1:N_l},\\
    F^{u}_{[ c ]} = \mathcal{F}(X^u_{[c]}), \quad
    F^{l}_{[ c ]} = \mathcal{F}(X^l_{[c]}),
\end{aligned}
\end{equation}
where $\mathcal{F}$ denotes the feature extractor, $X_{[c]}^{l}$ is the set of labeled samples with the ground-truth label equal to $c$, and $X_{[c]}^{u}$ is the set of unlabeled samples with the pseudo-label $\hat{y}$ equal to $c$. $F^{l}_{[c]}$ and $F^{u}_{[c]}$ are matrices that contain the extracted features of the labeled and unlabeled samples corresponding to category $c$, respectively. 

We can derive posterior class-wise GPs to make predictions about the unlabeled target samples as follows: 
\begin{equation}
\begin{aligned}
    \mathcal{GP}_c:\, F_{[c]}^{u}\,|\,F_{[c]}^{l},\mathcal{D}_l,\mathcal{D}_{ut}\sim\mathcal{N}(\mu_{u,c}, \Sigma_{u,c}),
\end{aligned}
\label{post_gp}
\end{equation}
with parameters
\begin{equation}
\begin{aligned}
    \mu_{u,c}\triangleq K(F_{[c]}^{u},F_{[c]}^{l})[K(F_{[c]}^{l},F_{[c]}^{l})]^{-1}F_{[c]}^{l},
\end{aligned}
\end{equation}
\begin{equation} 
\begin{aligned}
        \Sigma_{u,c}\triangleq K(F_{[c]}^{u},F_{[c]}^{u})-\\
\end{aligned}
\end{equation}
$$
\begin{aligned}
   K(F_{[c]}^{u},F_{[c]}^{l})[K(F_{[c]}^{l},F_{[c]}^{l})]^{-1}K(F_{[c]}^{l},F_{[c]}^{u}),    
\end{aligned}
$$
where $K$ is the linear kernel function defined as
\begin{equation}
    K(P,Q)_{j,k}=\frac{ P_jQ_k^T }{\left\| P_j \right\|\left\| Q_k \right\|}.
\end{equation}
Here $P_j$ is the $j$-th feature vector in $P$ and $Q_k$ is the $k$-th feature vector in $Q$. 

\medskip\noindent
\textbf{Query based on Posterior Variances (PV)}. For the $c$-th category, $\Sigma_{u,c}$ is the covariance matrix with the shape of $N_{u,c} \times N_{u,c}$ where $N_{u,c}$ is the number of unlabeled samples with the pseudo-label equal to $c$. We obtain the vector containing the predicted posterior variances for these samples as:
\begin{equation}
    PV_c=\mathrm{diag}(\Sigma_{u,c}),
    \label{post_var}
\end{equation}
where $\mathrm{diag}(\cdot)$ selects the diagonal elements of $\Sigma_{u,c}$, and $PV_c$ is of size $1 \times N_{u,c}$. We create a vector containing $PV$ values for all unlabeled target samples by concatenating $PV_c$ for all classes:
\begin{equation}
    PV=[PV_1\,  PV_2\,  ...  \,PV_C], \quad |PV|=N_u,
\end{equation}
where $N_u$ denotes the number of unlabeled samples. In each selection round, we select the top $b$ samples with the highest posterior variance from the sorted $PV$ and query their labels.

\subsection{Uncertainty-balanced Class Sampling}
Following the SENTRY algorithm \cite{prabhu2021sentry}, we incorporate unsupervised domain adaptation to align the target and source domains. SENTRY employs various data augmentations on target samples and uses their predicted pseudo-labels to identify consistent and inconsistent samples. It then minimizes the entropy of consistent images while maximizing that of inconsistent ones. Thus, the objective function of SENTRY is as follows:
\begin{equation}
    \mathcal{L}_{sent}=
    \begin{cases}
        -H(y|\tilde{x}^t_{ic}), & \text{if $x$ inconsistent}\\
        H(y|\tilde{x}^t_c), & \text{if $x$ consistent}\\
       
    \end{cases},
\end{equation}
where $\tilde{x}^t_{ic}$ is an inconsistent augmented image and $\tilde{x}^t_c$ is a consistent augmented image. $y$ denotes the output of the classifier and $H(\cdot)$ is the entropy function. During the adaptation process, a class-balanced sampling is performed on target samples. In each training epoch, the probability of selecting a sample is determined by the size of the class to which it belongs. However, balanced training only based on the class sizes leads to sub-optimal performance for ADA since it does not take the uncertainty of target samples into consideration.

Therefore, we propose Uncertainty-balanced Class Sampling (UCS) for target samples based on GP. Specifically, we compute the posterior variance of each unlabeled target sample using Eq. \ref{post_var}. We then calculate the average posterior variance for all target samples with a specific pseudo-label, i.e. class $c$, which provides the uncertainty measure of that particular class. These class-wise uncertainties are updated after each epoch using the Exponential Moving Average (EMA):
\begin{equation}
    U_n^{c} = \alpha U_{n-1}^{c} + (1-\alpha ) AV_{n}^{c},
    \label{ucs}
\end{equation}
where $U_n^{c}$ is the estimated uncertainty of class $c$ at epoch $n$, and $AV_{n}^{c}$ is the average posterior variance for all samples with the predicted class $c$. 

To construct the target training set at epoch $n$, we first assign a weight to each target sample based on the estimated uncertainty ($U_n^{c}$) corresponding to its pseudo-label. We then sequentially sample from these weighted samples until we reach the number of samples in the original target training set. It is noteworthy that we perform UCS after the GPAS and PLCS phases to ensure that the same sample is not selected multiple times during the sample selection process.

\medskip\noindent
\textbf{Overall Loss}.
We train the model by minimizing the cross-entropy loss on labeled data and the SENTRY loss on unlabeled data. The total objective function is given as follows: 
\begin{equation}
    \mathcal{L}_{total}=\mathcal{L}_{ce}(x_l,y)+\lambda \mathcal{L}_{sent}(x_{u}).
    \label{loss}
\end{equation}

\begin{table*}
    \centering
    \resizebox{0.96\textwidth}{!}{%
        \begin{tabular}{c|ccccccccccccc}
        \toprule
        \rowcolor{lightgray!20} \textbf{Method} & \multicolumn{13}{c}{\textbf{Office-Home}}\\
        \rowcolor{lightgray!20} & \textbf{A $\to$ C} & \textbf{A $\to$ P} & \textbf{A $\to$ R} & \textbf{C $\to$ A} & \textbf{C $\to$ P} & \textbf{C $\to$ R} & \textbf{P $\to$ A} & \textbf{P $\to$ C} & \textbf{P $\to$ R} & \textbf{R $\to$ A} & \textbf{R $\to$ C} & \textbf{R $\to$ P} & \textbf{Avg} \\
        \midrule
        ResNet~\cite{he2016deep} & 42.1 & 66.3 & 73.3 & 50.7 & 59.0 & 62.6 & 51.9 & 37.9 & 71.2 & 65.2 & 42.6 & 76.6 & 58.3\\
        \hline
        Random & 56.8 & 78.0 & 77.7 & 58.9 & 70.7 & 70.5 & 60.9 & 53.2 & 76.8 & 71.5 & 57.5 & 81.8 & 67.9\\
        Entropy & 56.8 & 80.0 & 82.0 & 59.4 & 75.8 & 73.8 & 62.3 & 54.6 & 80.3 & 73.6 & 58.8 & 85.7 & 70.2  \\
        CoreSet~\cite{sener2017active} & 51.8 & 72.6 & 75.9 & 58.3 & 68.5 & 70.1 & 58.8 & 48.8 & 75.2 & 69.0 & 52.7 & 80.0 & 65.1\\ 
        BADGE~\cite{ash2019deep} & 59.2 & 81.0 & 81.6 & 60.8 & 74.9 & 73.3 & 63.7 & 54.2 & 79.2 & 73.6 & 59.7 & 85.7 & 70.6\\
        \hline
        AADA~\cite{aada} & 56.6 & 78.1 & 79.0 & 58.5 & 73.7 & 71.0 & 60.1 & 53.1 & 77.0 & 70.6 & 57.0 & 84.5 & 68.3\\
        DBAL~\cite{de2021discrepancy}& 59.2 & 81.0 & 81.6 & 60.8 & 74.9 & 73.3 & 63.7 & 54.2 & 79.2 & 73.6 & 59.7 & 85.7 & 70.6\\
        CLUE~\cite{prabhu2021active}  & 58.0 & 79.3 & 80.9 & 68.8 & 77.5 & 76.7 & 66.3 & 57.9 & 81.4 & 75.6 & 60.8 & 86.3 & 72.5\\
        TQS~\cite{tqs}& 58.6 & 81.1 & 81.5 & 61.1 & 76.1 & 73.3 & 61.2 & 54.7 & 79.7 & 73.4 & 58.9 & 86.1 & 70.5\\
        SDM-AG~\cite{sdm}& 61.2 & 82.2 & 82.7 & 66.1 & 77.9 & 76.1 & 66.1 & 58.4 & 81.0 & 76.0 & 62.5 & 87.0 & 73.1\\
        \hdashline
        \textbf{Ours} & \textbf{66.0} & \textbf{83.8} & \textbf{83.8} & \textbf{69.1} & \textbf{79.1} & \textbf{79.5} & \textbf{69.5} & \textbf{65.6} & \textbf{84.8} & \textbf{76.5} & \textbf{69.6} & \textbf{87.3} & \textbf{76.2}\\
        \bottomrule
        \end{tabular}
    }
    \vspace{-2mm}
    \caption{Quantitative results on Office-Home. The classification accuracy (\%) is reported with an annotation budget of $5\%$.}
    \label{tab:officehome}
    \vspace{-2mm}
\end{table*}

\begin{table*}
    \renewcommand{\arraystretch}{1.1}
    \centering
    \resizebox{0.96\textwidth}{!}{%
        \begin{tabular}{c|ccccccccccccc}
        \toprule
        \rowcolor{lightgray!20} \textbf{Method} & \multicolumn{13}{c}{\textbf{DomainNet}}\\
        \rowcolor{lightgray!20} & \textbf{S $\to$ C} & \textbf{S $\to$ P} & \textbf{S $\to$ R} & \textbf{C $\to$ S} & \textbf{C $\to$ P} & \textbf{C $\to$ R} & \textbf{P $\to$ S} & \textbf{P $\to$ C} & \textbf{P $\to$ R} & \textbf{R $\to$ S} & \textbf{R $\to$ C} & \textbf{R $\to$ P} & \textbf{Avg}\\
        \midrule
        ResNet~\cite{he2016deep} & 63.0 & 66.5& 76.2&  60.5& 58.4 & 76.7& 66.2 & 61.1 & 82.6 & 58.4 & 65.8& 73.6 & 67.4\\
        \hline
        Random & 74.2 & 79.1 & 87.6& 73.2& 75.9& 87.3& 72.9& 73.1& 88.7& 70.2& 73.9& 78.8& 77.9\\
        Entropy & 78.1 & 81.1 & 89.3 & 76.5 & 78.3 & 91.0 & 74.7 & 74.6 & 90.5 & 72.1 & 77.0 & 81.5 & 80.4\\
        BADGE~\cite{ash2019deep}& 76.3& 82.9& 89.5& 76.3& 81.6& 91.1& 75.6& 72.9& 89.7& 72.3& 77.9& 79.4& 80.5\\
        \hline
        AADA~\cite{aada}& 74.6& 81.6& 91.0& 74.3& 74.3& 89.2& 73.0& 69.6& 89.2& 69.7& 77.6& 80.1& 78.7\\
        CLUE~\cite{prabhu2021active}& 76.2 & 77.9 & 88.8 & 74.5 & 76.8 & 88.5 & 74.2 & 68.3 & 89.5 & 70.4 & 72.7 & 79.1 & 78.2\\
        TQS~\cite{tqs} & 76.9& 80.6& 89.8& 76.4& 80.6& 90.7& 76.0& 74.8& 91.9& 74.4& 75.9& 82.6& 80.9\\
        SDM-AG~\cite{sdm} &  78.8& 82.5& 91.1& 79.5& 83.2& 91.9& 77.8& 78.2& \textbf{92.7}& 76.6& 78.3& 83.7& 82.9\\
        \hdashline
        \textbf{Ours} & \textbf{84.2} & \textbf{83.8} & \textbf{91.8} & \textbf{80.2} & \textbf{83.4} & \textbf{92.3} & \textbf{79.9} & \textbf{82.6} & 91.0 & \textbf{80.6} & \textbf{86.9} & \textbf{84.2} & \textbf{85.1}\\
        \bottomrule
        \end{tabular}
    }
    \vspace{-2mm}
    \caption{Quantitative results on DomainNet. The classification accuracy (\%) is reported with an annotation budget of $5\%$.}
    \label{tab:domainnet}
    \vspace{-4mm}
\end{table*}

\section{Experiments and Results}
We conduct extensive experiments on two commonly used DA datasets, namely DomainNet \cite{domainnet} and Office-Home \cite{officehome} to verify the effectiveness of our approach. We compare our method with existing ADA approaches, namely Source-Only (ResNet), CoreSet \cite{sener2017active}, AADA \cite{aada}, CLUE \cite{prabhu2021active}, TQS \cite{tqs}, DBAL \cite{de2021discrepancy}, and SDM-AG \cite{sdm}. We follow the protocol employed in prior studies and report the accuracy of each domain transfer for a dataset, as well as the average accuracy across all domain transfers. We perform five rounds of active sampling. In each round, we sample $1\%$ of the target data for annotation, resulting in a total budget of $5\%$.

\medskip \noindent
\textbf{Implementation details}. We use an ImageNet pre-trained ResNet-50 as the feature extractor and a fully connected layer as the classifier. Training is conducted using SGD with a learning rate of 0.002, momentum of 0.9, weight decay of 0.005, and a batch size of 16. The loss weight $\lambda$ is set to 1. Each sampling round is followed by 3 UDA epochs, with a five-epoch warm-up before the first round. For the PLCS phase, we initialize $\kappa = 1$ and increase it by 1 per round, resulting in $15\%$ certain pseudo-labels used during training.

\medskip \noindent
\textbf{Office-Home results}. The classification results for the Office-Home dataset are shown in Table \ref{tab:officehome}. This table shows the superiority of our GP-based approach in comparison to other uncertainty-based methods such as entropy-based sampling and TQS. Notably, our method outperforms TQS and ENT by significant margins of {$5.7\%$} and  {$6.0\%$}, respectively. Furthermore, the proposed method surpasses SDM-AG by  {$3.1\%$}. We observe that our method outperforms SDM-AG in more difficult domain transfers such as A$\to$C, P$\to$C, and R$\to$C by large margins of  {$4.8\%$}, {$7.2\%$} and   {$7.1\%$}, confirming the efficacy of our approach to adapt to challenging domains.

\medskip \noindent
\textbf{DomainNet results}. The results of our proposed method on the DomainNet dataset show a significant improvement compared to other state-of-the-art ADA and AL algorithms. Table \ref{tab:domainnet} shows the performance comparison of our method with SDM-AG, CLUE, and AADA. Our approach achieves an average accuracy of  {$85.1\%$} on DomainNet, which is a substantial improvement over the other methods. Specifically, our method outperforms SDM-AG by  {$2.2\%$}, CLUE by  {$6.9\%$}, and AADA by  {$6.4\%$} on average.

\begin{figure*}[t!]
    \begin{center}
        \includegraphics[width=0.80\linewidth]{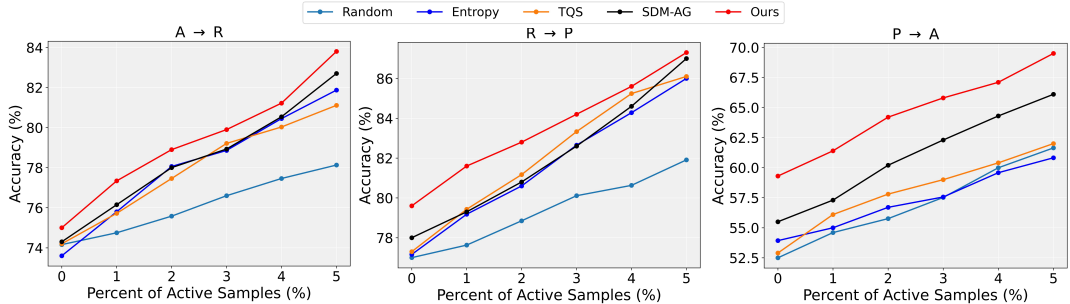}
    \end{center}
    \vskip -20.0pt
    \caption{Comparison between different ADA approaches over AL rounds on three different domain transfers of Office-Home.}
    \label{fig:vary_active_gp} 
    \vskip -5.0pt
\end{figure*}

\section{Ablation Study}
\noindent
\textbf{Effect of different components}.
We conduct extensive ablation studies to assess the effectiveness of each component in our method, reporting results in Table \ref{tab:ablation} on the Office-Home dataset. The first row presents unsupervised domain adaptation using SENTRY. Variant 1 applies random sampling with SENTRY loss, while Variant 2 uses only GPAS for uncertain sample selection, yielding a notable $3.3\%$ improvement over random sampling. Variant 3 further integrates UCS, sampling based on target class uncertainties (Eq. \ref{ucs}), boosting performance by $0.5\%$. Finally, combining GPAS, PLCS, and UCS achieves a classification accuracy of $76.2\%$, demonstrating that each component significantly enhances performance.



\begin{table}
    \renewcommand{\arraystretch}{1.1}
    \centering
    \resizebox{0.40\textwidth}{!}{%
    \setlength{\tabcolsep}{1mm}
    \begin{tabular}{c|c|ccc|c}
    \toprule
    \rowcolor{lightgray!20} \textbf{Method} & \textbf{UDA} & \multicolumn{3}{c|}{\textbf{Active Sampling}} & \textbf{Avg. Acc.} \\ \cline{3-5}
    \rowcolor{lightgray!20} & \textbf{with UCS} & \textbf{Random} & \textbf{GPAS} & \textbf{PLCS} & \\ \hline
    UDA          & -          & -          & -          & -          & 69.6  \\ \hline
    Variant 1    & -          & \checkmark & -          & -          & 71.6  \\  
    Variant 2    & -          & -          & \checkmark & -          & 74.9  \\  
    Variant 3    & \checkmark & -          & \checkmark & -          & 75.4  \\ \hline
    \textbf{Ours} & \checkmark & -          & \checkmark & \checkmark & 76.2  \\ 
    \bottomrule
    \end{tabular}}
    \caption{Ablation study for our approach on Office-Home.}
    \label{tab:ablation}
    \vskip -10pt
\end{table}

\noindent


\begin{table}[!t]
    \renewcommand{\arraystretch}{0.95}
    \centering
    \resizebox{0.30\textwidth}{!}{%
    \setlength{\tabcolsep}{1mm}
    \begin{tabular}{c|cc|c}
    \toprule
    \rowcolor{lightgray!20} \textbf{Method} & \textbf{PLCS} & \textbf{SENTRY} & \textbf{Avg. Accuracy} \\ 
    \midrule
    \multirow{3}{*}{Random}  &      -  &     - &   67.9   \\ 
                             &          \checkmark  &    -  &    69.1  \\ 
                             &          \checkmark     &   \checkmark   &  72.4    \\ \hline
    \multirow{3}{*}{Entropy} &     -   &   -   &   70.2   \\ 
                             &           \checkmark    & -   &   71.2   \\ 
                             &             \checkmark  & \checkmark   &   74.2   \\ \hline
    \multirow{3}{*}{CLUE \cite{prabhu2021active}}    &   -          &   - & 72.5 \\  
                             &            \checkmark   &   -   &   73.2 \\
                             &          \checkmark    &  \checkmark    &    74.7  \\ \hline
    
    \multirow{3}{*}{SDM-AG \cite{sdm}}  &         -    &   -   &    73.1  \\
                             &            \checkmark   &   -   &     74.4 \\
                             &          \checkmark    &   \checkmark   &    75.4  \\ \hline
    \textbf{GP (ours)}                         &    \checkmark    &   \checkmark   &    \textbf{76.2}  \\ \bottomrule
    \end{tabular}}  

\caption{Effect of integrating PLCS and SENTRY loss into other ADA methods (Office-Home).}
\label{ablation11}
\vskip -6.0pt
\end{table}

\medskip\noindent
\textbf{GPAS versus other ADA methods}. To quantify uncertainty, our GPAS strategy models the joint distribution
of both source and target samples, taking into consideration the relative effect of these samples on each other. To verify the effectiveness of our uncertainty measure, in Table \ref{ablation11}, we conduct an ablation study where we replace our GPAS active sampling algorithm with other SOTA ADA methods while keeping the SENTRY and PLCS components. Specifically, for each compared method in Table \ref{ablation11}, the first row shows the original performance on the Office-Home dataset, the second row indicates the performance with PLCS integrated, and the third row displays the performance with both SENTRY loss and PLCS integrated into the method. From the results shown in the third row for each method, we can observe that our approach outperforms all baselines which shows the effectiveness of the proposed GPAS strategy.


\medskip\noindent
\textbf{Integrating PLCS to other methods}. Table \ref{ablation11} evaluates the impact of our PLCS algorithm on various ADA methods. We test on the Office-Home dataset, comparing random sampling, entropy sampling, SDM-AG, and CLUE, with and without PLCS. The results show that integrating PLCS improves ADA performance by reducing the search space. Notably, SDM-AG and CLUE achieve $1.3\%$ and $0.7\%$ higher DA performance, respectively, with PLCS. This demonstrates that PLCS enhances both ADA and standard AL methods.

\medskip\noindent
\textbf{Performance Improvement over Rounds}. Figure \ref{fig:vary_active_gp} presents accuracy curves for various ADA algorithms over five active sampling rounds. Our method consistently outperforms others in each round, demonstrating its effectiveness. The total pseudo-label sampling rate is set to $15\%$.

\medskip\noindent
\textbf{Execution efficiency.} Table \ref{tab:time_complexity} compares the AL query time per round of our GP-based active selection with CLUE and Entropy on Office-Home (A $\to$ C) and DomainNet (R $\to$ P). The results show that our query function is as time-efficient as other baselines while achieving superior performance.

\begin{table}[h]
    \setlength{\tabcolsep}{4pt}
    \centering
    \resizebox{.6\columnwidth}{!}{%
        \begin{tabular}{c c c} 
        \toprule
        \rowcolor{lightgray!20} \multirow{2}{*}{\textbf{Method}} & \textbf{Office-Home} & \textbf{DomainNet} \\
        \cmidrule(lr){2-3}
        \rowcolor{lightgray!20} & \textbf{A $\to$ C} & \textbf{R $\to$ P} \\
        \midrule
        Entropy & 81.6s & 151.4s  \\
        CLUE & 93.5s & 161.5s \\
        \emph{Ours} & 89.0s & 133.3s \\
        \bottomrule
        \end{tabular}%
    }
    \caption{Query time complexity of ADA approaches.}
    \label{tab:time_complexity}
\end{table}

\vspace{-6mm}
\section{Conclusion}
We proposed a novel collaborative framework for Active Domain Adaptation (ADA) that incorporates both uncertain and certain target samples during the training process. Our approach involves using a Gaussian Process-based Active Sampling (GPAS) strategy to identify uncertain samples and a Pseudo-Label-based Certain Sampling (PLCS) strategy to identify highly confident samples. By incorporating these two strategies, our approach significantly reduces the active sampling search space and boosts the certain sample rate, leading to better adaptation and improved performance on several domain adaptation datasets. Overall, our framework provides a more effective approach to ADA, allowing for better adaptation to the target domain by integrating both uncertain and certain samples into the training process.

\bibliographystyle{IEEEbib}
\bibliography{refs}

\clearpage
\section{Appendix}

\subsection{Varying Certain Sampling Rate}
\begin{figure}[t!]
    \begin{center}
        \includegraphics[width=0.60\linewidth]{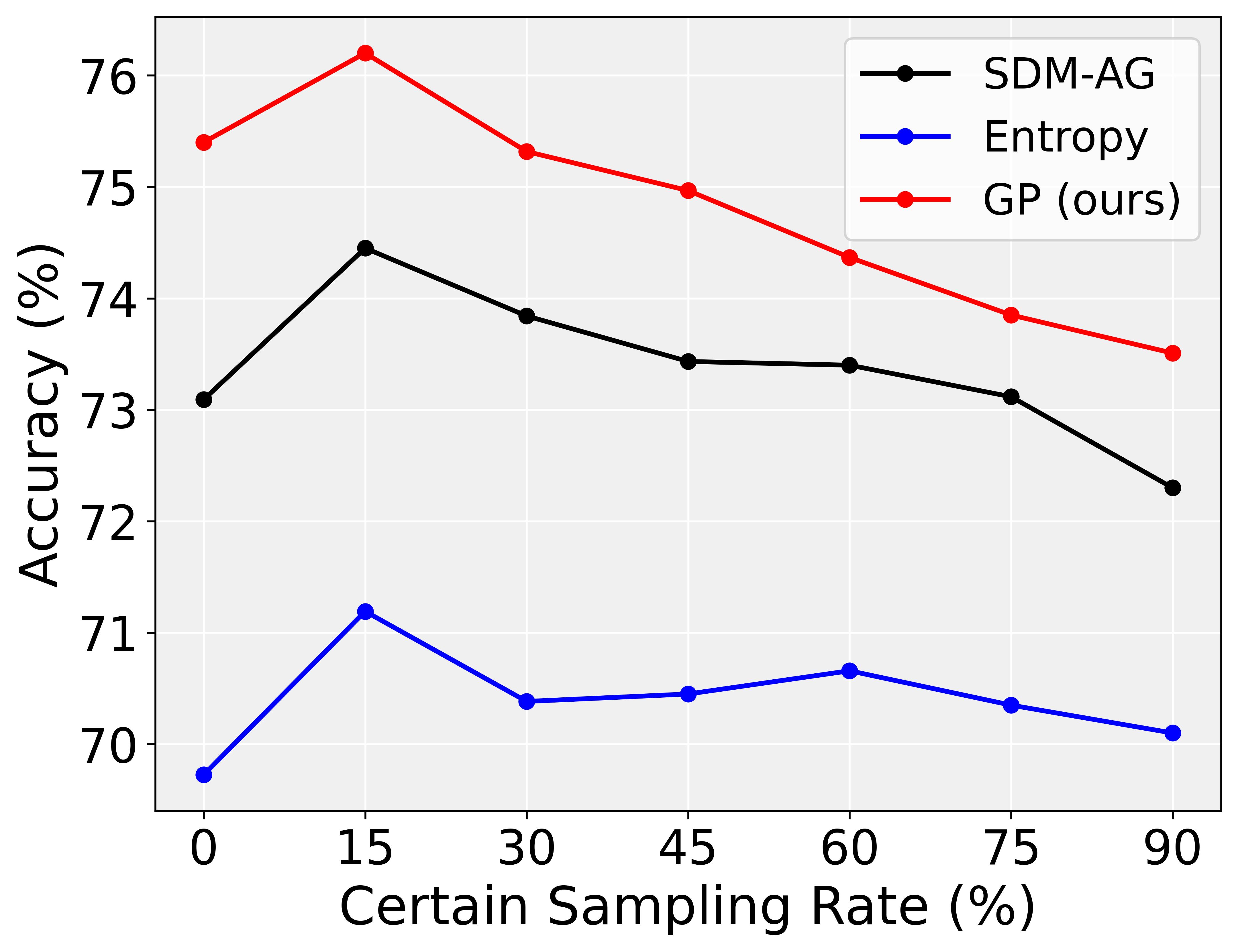}
    \end{center}
    \vskip -10.0pt
    \caption{Comparison between our method, SDM-AG, and entropy sampling with fixed $5\%$ active sampling rate while varying certain sampling rate from $0\%$ to $90\%$ on the Office-Home dataset.}
    \label{fig:vary_pl_gp} 
    \vskip -5.0pt
\end{figure}

Fig. \ref{fig:vary_pl_gp} illustrates the results of experiments conducted to observe the impact of varying the certain sampling rate on the performance of our method, SDM-AG, and entropy sampling approaches. The experiments were carried out by fixing the active sampling rate at $15\%$ and varying the certain sampling rate from $0\%$ to $90\%$. We observe that increasing the rate of pseudo-labeling initially contributes to improved model performance. However, beyond a certain threshold, the number of false-positive pseudo-labels begins to increase, which has a negative impact on the model's performance. We observed that the value of $15\%$ for the certain sampling rate in our experiments consistently obtained optimal performance across all DA datasets. 

\subsection{GPAS Versus Other ADA Methods}
To quantify uncertainty, our GPAS strategy models the joint distribution
of both source and target samples, taking into consideration the relative effect of these samples on each other. To verify the effectiveness of our uncertainty measure, in Table \ref{ablation11}, we conduct an ablation study where we replace our GPAS active sampling algorithm with other SOTA ADA methods while keeping the SENTRY and PLCS components. Specifically, for each compared method in Table \ref{ablation11}, the first row shows the original performance on the Office-Home dataset, the second row indicates the performance with PLCS integrated, and the third row displays the performance with both SENTRY loss and PLCS integrated into the method. From the results shown in the third row for each method, we can observe that our approach outperforms all baselines which shows the effectiveness of the proposed GPAS strategy.

\subsection{Varying Total Budget of Active Sampling} 
In Table \ref{budget_increase}, we present an experiment conducted on the Office-Home dataset to examine the effect of varying the total active sampling budget on the DA performance using our method. In this experiment, we set the total certain sampling rate of our method at a fixed value of $15\%$ while varying the total active sampling budget $\mathcal{B}$ from $0\%$ to $20\%$. As shown in Table \ref{budget_increase}, we observe that the performance improvement starts to diminish as we increase the total budget. This indicates that our proposed method is capable of selecting the most uncertain and useful target samples even within a low budget. Consequently, the samples chosen under higher budgets provide less informative value to the model, resulting in less performance improvement.
\begin{table}
\renewcommand{\arraystretch}{1.1}
    \centering
    \resizebox{0.47\textwidth}{!}{$
    \setlength{\tabcolsep}{1mm}{
    \begin{tabular}{c|c|c|c|c|c}
    \hline
    Total Budget of Sample Selection ($\mathcal{B}$)  & $0\%$ & $5\%$ & $10\%$ & $15\%$ & $20\%$ \\ \hline
    Avg. Acc.     &  69.6  &  76.2  & 79.3   &  81.2  &  83.0  \\ \hline
    Improvement &  -  &  6.6  &  3.1  &  1.9  &  1.8   \\ \hline
    \end{tabular}}
    $}

\caption{Effect of increasing the total budget of sample selection on the classification accuracy for the Office-Home dataset.}
\label{budget_increase}
\end{table}

\subsection{Additional Implementation Details}

Our method is implemented in PyTorch \cite{NEURIPS2019_9015}. For all experiments, we use an ImageNet \cite{deng2009imagenet} pre-trained ResNet-50 \cite{he2016deep} architecture as our feature extractor and a single fully-connected layer as the classifier. For the classifier, we initialize weights using the Xavier initialization technique with no bias. We use SGD optimizer \cite{ruder2016overview} with a learning rate of 0.002, a momentum of 0.9, and a weight decay of 0.005. Our learning rate scheduler is similar to the one used in \cite{long2018conditional}. We use a batch size of 16 for all experiments. The $\lambda$ is set to 1 for loss calculation. For the results on DomainNet, we only utilized the training set of both source and target domains in our experiments and excluded the test sets. For image preprocessing, we first resize images to 256 pixels, then we apply RandomCrop of size 224 pixels followed by RandomHorizontalFlip during the training process. During the testing process, we only resize the images to 224 pixels. After each sampling round, we generally perform three UDA epochs, and we warm up our model for five UDA epochs prior to the first sampling round. For the PLCS phase, we generally adopt an initial $\kappa$ of 1 and we increase it in each sampling round by 1. Hence, we use $15\%$ of certain pseudo-labels during the training process. We utilize an NVIDIA A5000 GPU to run our experiments.

\subsection{Baselines}
We compare the proposed method with several existing ADA approaches, namely Source-Only (ResNet), Random, Entropy, CoreSet \cite{sener2017active}, AADA \cite{aada}, CLUE \cite{prabhu2021active}, TQS \cite{tqs}, DBAL \cite{de2021discrepancy}, and SDM-AG \cite{sdm}.

\noindent
\textbf{Source-Only}: In Source-Only, we report the target accuracy of a ResNet model only trained on the source data.
\par\vspace{1mm}
\noindent
\textbf{Random}: In random sampling, target samples are randomly selected for annotation.
\par\vspace{1mm}
\noindent
\textbf{Entropy}: Entropy-based sampling selects samples based on the model's predictive entropy.
\par\vspace{1mm}
\noindent
\textbf{CoreSet}: CoreSet is an AL algorithm that relies on the concept of core-set selection.
\par\vspace{1mm}
\noindent
\textbf{AADA}: In AADA, target samples are chosen using both predictive entropy and targetness criteria.
\par\vspace{1mm}
\noindent
\textbf{CLUE}: CLUE computes the uncertainty of target samples and then performs an uncertainty-weighted clustering method in order to ensure diverse sampling.
\par\vspace{1mm}
\noindent
\textbf{TQS}: TQS employs multiple classifiers and finds the samples with the highest level of disagreement among these classifiers.
\par\vspace{1mm}
\noindent
\textbf{DBAL}: DBAL performs active sampling using a discrepancy-based strategy.
\par\vspace{1mm}
\noindent
\textbf{SDM-AG}: SDM-AG trains the model using a margin loss and proposes a variant of margin sampling for sample selection.

\subsection{Related Work}
\noindent
\textbf{Active Learning (AL)}. The objective of AL \cite{ash2019deep, hacohen2022active, safaei2025filter, safaei2025active, bang2024active, xie2023active} is to maximize the performance gain of a model by selecting the most useful samples from unlabeled data, annotating them, and integrating them into the supervised training process. Existing AL methods are generally classified into uncertainty-based and diversity-based methods. Uncertainty-based algorithms use different metrics, such as entropy \cite{luo2013latent, safaei2024entropic}, margin of confidence \cite{balcan2007margin}, and mutual information \cite{kirsch2019batchbald} to select uncertain samples. Diversity-based \cite{sener2017active,xu2003representative} strategies attempt to cluster the samples in the feature space and select a diverse set of samples by picking from different clusters. BADGE \cite{ash2019deep} combines both diversity and uncertainty to achieve improved performance by utilizing the magnitude of the model's gradients as a measure of uncertainty. Furthermore, in \cite{wei2015submodularity} uncertainty sampling and core-set selection concepts are combined through submodular functions. However, these general-purpose AL algorithms perform suboptimally when the labeled and unlabeled data are sampled from different distributions.

\noindent
\textbf{Domain Adaptation (DA)}. Domain adaptation aims at transferring the knowledge of models from a labeled source domain to a target domain where target annotations are not available (UDA) \cite{ganin2015unsupervised,hoffman2018cycada} or are scarce (semi-supervised DA) \cite{saito2019semi,li2021learning}. Many methods attempt to align the distribution of source and target domains in the latent space by minimizing the domain misalignment statistics \cite{kang2019contrastive,yan2017mind}. Another set of works employs domain-adversarial training to produce domain-invariant features \cite{ganin2016domain,gao2021gradient}. Recently, self-training approaches have become popular in UDA where they generally rely on the confident predictions of target samples \cite{zou2018unsupervised,tan2020class} or regularizing their confidence \cite{zou2019confidence} to train the model. Another line of work is consistency-based UDA under different data augmentations \cite{prabhu2021sentry,bahat2019natural}. SENTRY \cite{prabhu2021sentry} enhances model confidence in consistent images by minimizing predictive entropy while maximizing predictive entropy on inconsistent images. 

\noindent
\textbf{Active Domain Adaptation (ADA)}. To address the shortcomings of general-purpose AL methods for domain adaptation, the problem of ADA was first introduced in \cite{rai2010domain}. AADA \cite{aada} selects samples that have high predictive entropy and targetness using a domain discriminator and performs adversarial domain alignment using DANN \cite{ganin2016domain}. CLUE \cite{prabhu2021active} uses predictive entropy for uncertainty estimation and then samples from different clusters that are weighted by entropy to impose diversity. However, these methods do not consider the relationship between unlabeled target samples, which may lead to redundant sampling. In our work, the GP models the joint distribution of both source and target samples, taking into consideration the relative effect of target samples on each other. TQS \cite{tqs} and S3VAADA \cite{rangwani2021s3vaada} combine multiple query functions which results in complex sample selection strategies overfitting to a particular domain shift. EADA \cite{xie2022active} and SDM-AG \cite{sdm} are loss-based approaches that aim to reduce the domain gap by introducing an auxiliary loss and selecting informative samples based on the scores from those losses. In \cite{de2021discrepancy}, the authors propose an ADA method based on the discrepancy of source and target distributions. Existing ADA algorithms do not use confident pseudo-labels of target samples during the training process. Our work introduces a combination of certain and uncertain sampling, which effectively boosts the adaptation performance.

\subsection{Algorithm}
Our proposed algorithm for active domain adaptation can be seen in Algorithm \ref{algo:meta}.
\begin{algorithm}[tb]
\caption{Our Proposed Algorithm for ADA}
\begin{algorithmic}[1]
{\footnotesize
\REQUIRE Labeled source data $\mathcal{D}_s$, unlabeled target data $\mathcal{D}_{ut}$, number of epochs $N$, active sampling rounds $R$, PLCS parameters $\kappa_{start}$ and $\kappa_{step}$, feature extractor $\mathcal{F}$ and classifier $\mathcal{C}$, per-round selection budget $b$
\STATE $\mathcal{D}_l \gets \mathcal{D}_s$ \quad \emph{{\# Labeled data}}
\STATE $\mathcal{D}_{lt} \gets \phi$ \quad \emph{{\#Labeled target data}}
\FOR{$n = 1$ to $N$}
    \IF{$n$ in $R$} 
      \STATE \emph{{\# \textbf{PLCS Strategy}}}
      \STATE $\forall x \in D_{ut}$, compute $\hat{y} 
      (x)\gets \arg\max\mathcal{C}(\mathcal{F}(x))$ \label{line:repeat}
      \STATE For the $i$-th class: \\ Conf$[i]\gets \left\{  \max\mathcal{C}(\mathcal{F}(x)) |x\in D_{ut},\,\hat{y}(x)=i\right\}$
      \STATE For the $i$-th class: $X_{conf}[i]\gets$ select the top $\kappa_{start}\%$ from the sorted Conf$[i]$,
      \STATE $\mathcal{D}_{lt} \gets \mathcal{D}_{lt} \cup X_{conf}[i]$,
      \STATE  $\mathcal{D}_{ut} \gets \mathcal{D}_{ut} \setminus X_{conf}[i]$.
      \STATE $\kappa_{start} = \kappa_{start} + \kappa_{step}$.
      \STATE \emph{{\# \textbf{GPAS Strategy}}}
      \STATE Repeat line \ref{line:repeat} to get pseudo-labels.
      \STATE Construct posterior class-wise GPs based on the obtained pseudo-labels using Eq. \ref{post_gp},
      \STATE $PV \gets \bm{\phi}$
      \STATE For the $i$-th class: compute posterior variance vector $PV_i$ using Eq. \ref{post_var},
      \STATE $PV$.append($PV_i$).
      \STATE $S \gets$ select the top $b$ samples from the sorted $PV$, and annotate them.
      \STATE $\mathcal{D}_{lt} \gets \mathcal{D}_{lt} \cup S$,
      \STATE  $\mathcal{D}_{ut} \gets \mathcal{D}_{ut} \setminus S$
      \STATE $\mathcal{D}_{l} \gets \mathcal{D}_{l} \cup \mathcal{D}_{lt}$.

    \ENDIF
    \STATE \emph{{\# \textbf{UDA}}}
    \STATE Construct an uncertainty-balanced data loader for target samples using Eq. \ref{ucs}.
    \STATE Minimize $\mathcal{L}_{total}$ as defined in Eq. \ref{loss}
 \ENDFOR
 
}
\end{algorithmic}
\label{algo:meta}
\end{algorithm}

\end{document}